\def\year{2020}\relax
\documentclass[letterpaper]{article} 
\usepackage{aaai20}  
\usepackage{times}  
\usepackage{helvet} 
\usepackage{courier}  
\usepackage[hyphens]{url}  
\usepackage{graphicx} 
\usepackage{graphicx}  
\usepackage{CJKutf8}
\usepackage{url}
\usepackage{subcaption}
\usepackage{multirow}
\usepackage{multibib}
\usepackage{amsfonts}
\usepackage{amsmath}
\usepackage{graphicx}
\usepackage{algorithm}
\usepackage{algorithmic}
\newcommand{\citet}[1]{\citeauthor{#1} \shortcite{#1}}
\newcommand{\citep}{\cite}

\title{Generating Diverse Translation by Manipulating Multi-Head Attention}
\author{
Zewei Sun,
Shujian Huang,
Hao-Ran Wei,
Xin-yu Dai,
Jiajun Chen\\
State Key Laboratory for Novel Software Technology \\
Nanjing University, Nanjing 210023, China \\
sunzw@smail.nju.edu.cn, whr94621@foxmail.com\\
\{huangsj, daixinyu, chenjj\}@nju.edu.cn
}
\begin{document}

\maketitle

\begin{abstract}
Transformer model \cite{Vaswani2017AttentionIA} has been widely used in machine translation tasks and obtained state-of-the-art results. In this paper, we report an interesting phenomenon in its encoder-decoder multi-head attention: different attention heads of the final decoder layer align to different word translation candidates. We empirically verify this discovery and propose a method to generate diverse translations by manipulating heads. Furthermore, we make use of these diverse translations with the back-translation technique for better data augmentation. Experiment results show that our method generates diverse translations without a severe drop in translation quality. Experiments also show that back-translation with these diverse translations could bring a significant improvement in performance on translation tasks. An auxiliary experiment of conversation response generation task proves the effect of diversity as well. 
\end{abstract}

\section{Introduction}

In recent years, neural machine translation (NMT) has shown its ability to produce precise and fluent translations \cite{Sutskever2014SequenceTS,Bahdanau2015attention,Luong2015EffectiveAT}. More and more novel network structures has been proposed \cite{Barone2017DeepAF,Gehring2017ConvolutionalST,Vaswani2017AttentionIA}, among which Transformer \cite{Vaswani2017AttentionIA} achieves the best results.
The main differences between Transformer and other translation models are: i) self-attention architecture, ii) multi-head attention mechanism. We focus on the second one in this paper. 

Intuitively, the attention mechanism in traditional attention-based sequence-to-sequence models plays the role of choosing the next source word to be translated, which could be seen as an alignment between source and target words \cite{Bahdanau2015attention,Luong2015EffectiveAT}. However, how multi-head attention works seems unclear.


In this paper, we report an interesting phenomenon in Transformer: in the final layer of its decoder, each individual encoder-decoder attention head dispersedly aligns to a specific source word which is highly likely to be translated next. In other words, multi-head attention actually learns multiple alignment choices. Further, by means of picking different attention heads, we can precisely control the following word generation. We verify this characteristic by a series of statistic study afterwards. 






Straightway, we consider taking advantage of this intrinsic characteristic to generate diverse translations due to the multiple generation candidates. Natural language can be diversely translated through different syntax structures or word orders. However, it has been well recognized that NMT system severely lacks translation diversity, distinguished from human beings \cite{He2018SequenceTS,Ott2018AnalyzingUI,Edunov2018UnderstandingBA}. We try to tackle this issue with a new method based on our observation.


There have been a few works attempting to generate more diverse translation, which can be roughly divided into two categories. The first category tries to encourage diversity during beam search by adding some regularization items \cite{Li2016ASF,Vijayakumar2018DiverseBS}. However, these methods actually fail to produce satisfied diversity in our re-implementation experiments. The other category tends to augment diversity by introducing latent variables \cite{He2018SequenceTS,Shen2019MixtureMF}. However, they heavily increase training barrier and lack interpretability.

Different from them, we make use of the diverse factors we observe inside the model structure, which is more lightweight and interpretable. Since multi-head attention has the potential to identify different translation candidates, we propose a method to manipulate it to generate diverse translations. Our method works simply but more effectively than previous works, bringing in no extra parameter or regularization item.
Furthermore, we propose to combine diverse translations with the back-translation technique for better data augmentation. 
  
Experiment results show that the proposed method could generate diverse translations without a severe drop in translation performance. Besides, improvements could be achieved by employing our more diverse back-translation results for machine translation. An auxiliary experiment of conversation response generation proves the effect of diversity as well.

\section{Background}

Transformer \cite{Vaswani2017AttentionIA} architecture adopts the encoder-decoder structure, utilizing self-attention instead of recurrent or convolutional networks.
The encoder iteratively processes its hidden representation through 6 layers of self-attention and feed-forward network, coupled with layer normalization and residual connection. Afterwards, the decoder takes a similar circuit with injected by an encoder-decoder attention layer between self-attention and feed-forward components. An important difference from previous models is that Transformer turns all the attention mechanism into a multi-head version.

\begin{figure}
    \centering
    \includegraphics[scale=0.25]{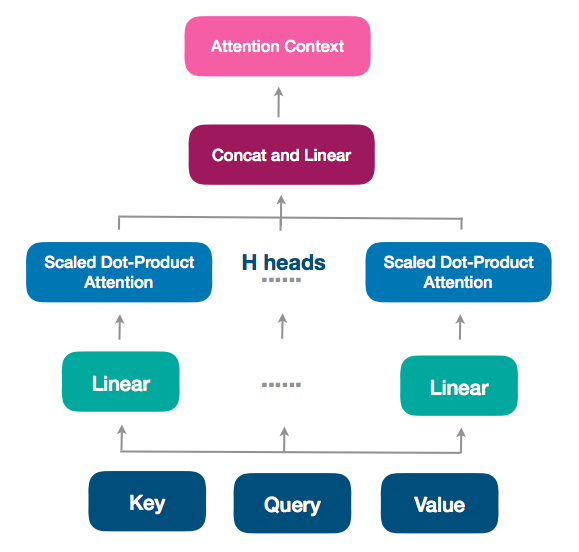}
    \caption{Multi-Head Attention consists of several attention heads running in parallel.}
    \label{fig:multi-head}
\end{figure}

Instead of performing a single attention function with $d$-dimensional keys, values and queries, multi-head attention projects them into $H$ different sub-components. After calculating attention for every sub-component, each yielding a $d/H$-dimensional
output context, these context vectors are concatenated and projected, resulting in the final context, as depicted in Figure~\ref{fig:multi-head}. Specifically:
\begin{equation}
     \text{MultiHead}(Q,K,V) = \text{Concat}(\text{head}_\text{1}, ... , \text{head}_\text{h})W^o
\end{equation}
\begin{equation}
    \text{head}_\text{i} = \text{Attention}(QW_i^Q, KW_i^K, VW_i^V)
\end{equation}
\begin{equation}
    \text{Attention}(Q,K,V) = \text{softmax}(\frac{QK^T}{\sqrt{d_k}})V
\end{equation}

To complete the decoding part, the model uses learned linear transformation and softmax function to convert the decoder output to next-token probabilities. The embedding and final transformation parameters are shared mutually.

\section{Analysis of Multi-Head Attention}
\label{sec:phenomenon}

\begin{figure}[b]
    \centering                 
    \includegraphics[scale=0.25]{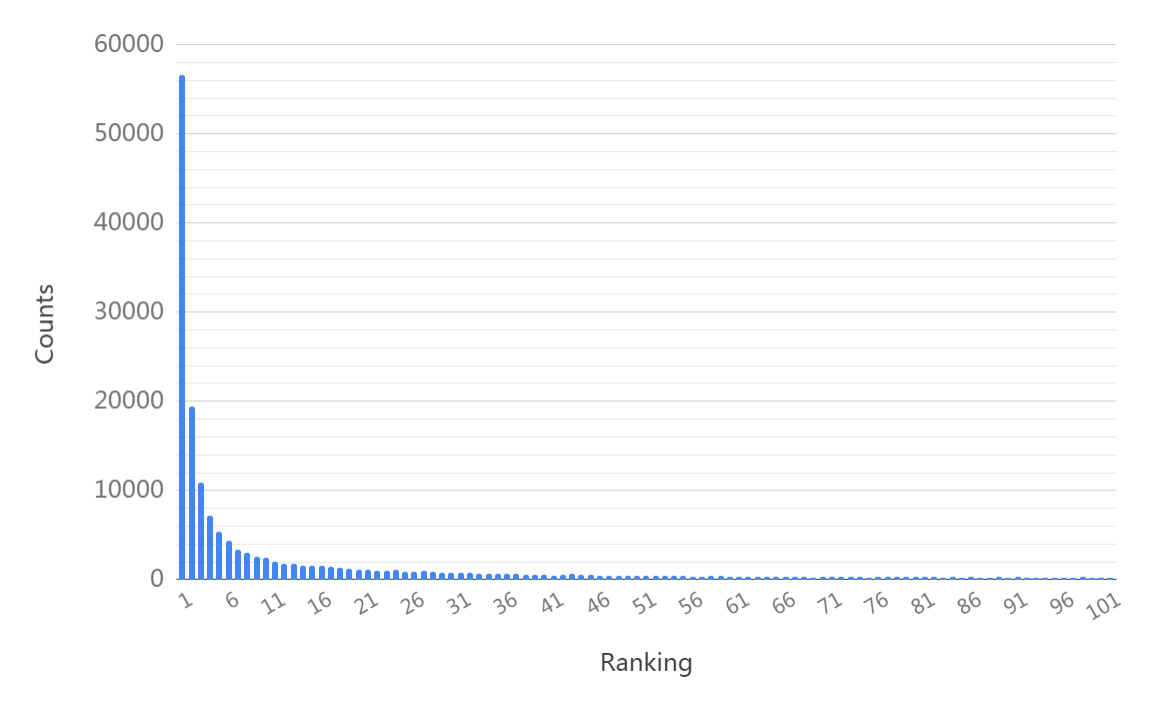}   
    \caption{``Referred target words" ranking counts of top 100. The vast majority of heads align to the most possible words.}
    \label{fig:rank-number}
\end{figure}

\begin{CJK}{UTF8}{gbsn}
    \begin{figure*}
        \centering                 
        \includegraphics[scale=0.16]{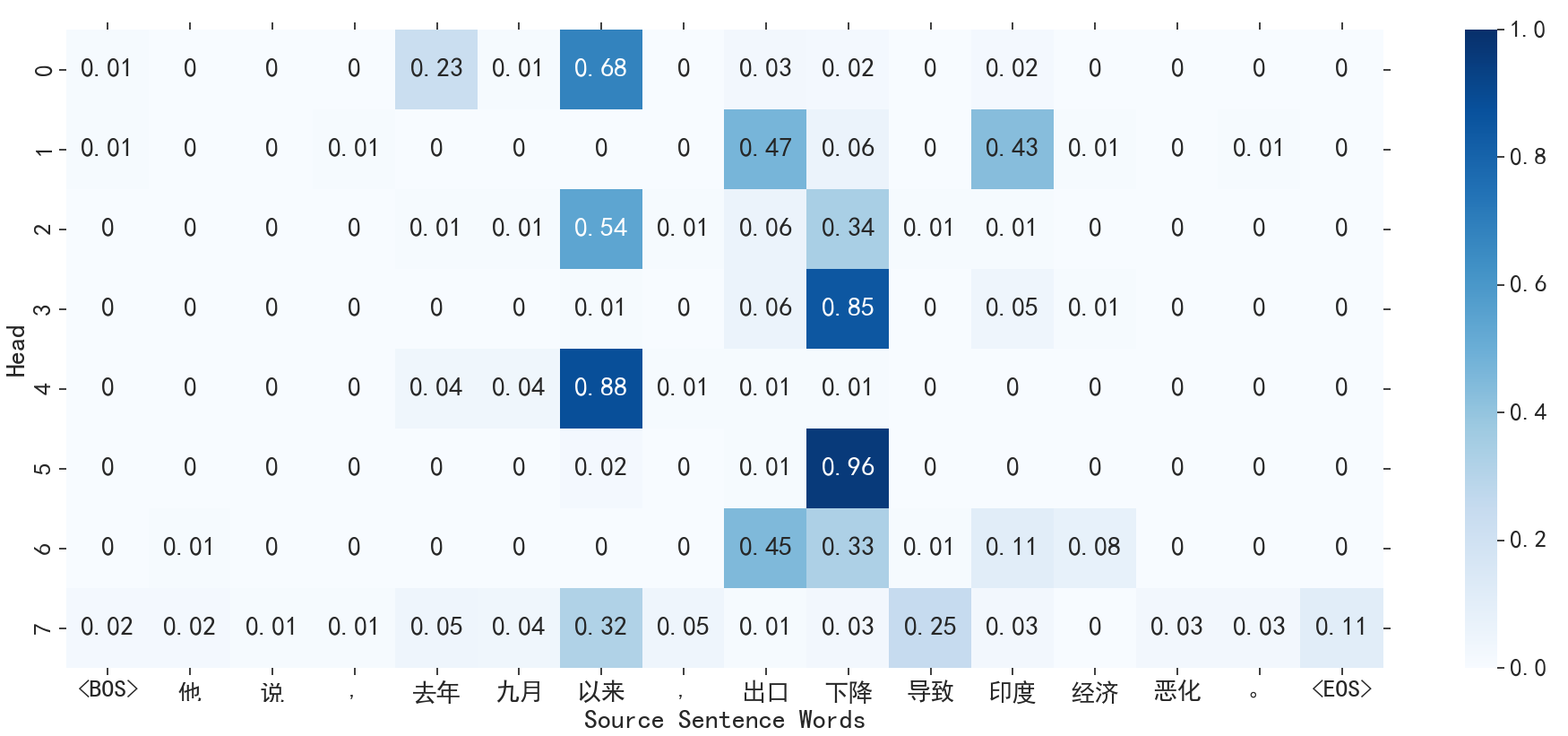}   
        \caption{Different heads have different attention, referring to different words. For example, head 4,5,6 refer to 以来 (since), 下降 (decline), 出口 (exports), respectively.}
        \label{fig:attention-case}
    \end{figure*}
\end{CJK}

\subsection{Each Head Indicates an Alignment}
Previous works show that multi-head attention plays a key role in the significant improvement of translation performance
\cite{Vaswani2017AttentionIA,Chen2018TheBO}. 
However, not much observation was made on its inside pattern. We visualize the multi-head attention to see whether different heads play different roles. After observing plenty of examples, we find that at every decoding step, all the source words that are identified by heads are highly likely to be translated next. In other words, each head aligns to a source word candidate. 


We do the statistic study to verify this observation on NIST MT03 dataset (see Datasets in Experiment). At each timestep, we pick $H$ referred source words (may overlap) that $H$ heads correspond to. ``Referred source word" means the source word with max attention for each head. Then we translate these referred source words into target language with baseline model and name these translations ``referred target words''. We count the number of times these ``referred target words'' appear at different rankings of the softmax probability and plot them in Figure~\ref{fig:rank-number}. We can see that the vast majority of heads align to the most possible words.

Also, we collect the negative log-likelihood (NLL) of these referred target words to see whether they really have high generation probability. To make a comparison, we list the average NLL of words ranked $R^{th}$ as well. The results in Table~\ref{table:nll} verify our assumption. The chosen words are ranked around $5^{th}$ on average, which implies they are indeed quite possible to be selected at each decoding step.

\begin{table}\footnotesize
    \centering
    \begin{subtable}{.2\textwidth}
        \centering
        \begin{tabular}{lc}
             & NLL \\ \hline
            Rank 1 & 0.59 \\
            Rank 3 & 4.12 \\
            Rank 5 & 5.19 \\
            \textbf{Head-Average} & \textbf{5.27} \\
            Rank 6 & 5.53 \\
            Rank 10 & 6.40 \\
            Rank 100 & 9.47 \\
            Rank 1000 & 11.54 \\
            Random & 13.57 
        \end{tabular}
    \end{subtable}
    \begin{subtable}{.2\textwidth}
        \centering
        \begin{tabular}{lc}
             & NLL \\ \hline
            Head 0 & 5.05 \\
            Head 1 & 5.40 \\
            Head 2 & 4.69 \\
            Head 3 & 5.04 \\
            Head 4 & 5.33 \\
            Head 5 & 5.72 \\
            Head 6 & 5.68 \\
            Head 7 & 5.32 \\
            Head-Average & 5.27
        \end{tabular}
    \end{subtable}
    \caption{Negative log-likelihood of attention heads and words ranked $R^{th}$ on average.}
    \label{table:nll}
\end{table}

\subsection{Each Head Determines a Generation Word}
\begin{CJK}{UTF8}{gbsn}
Furthermore, we can control the next word generation by choosing the corresponding source word by choosing different heads. As presented in Table~\ref{table:controlling}, the model has translated ``he said :" and waits for the following context. 
At this step, different heads refer to several source words. From Figure~\ref{fig:attention-case}, we can see that head 4,5,6 refer to 以来 (since), 下降 (decline), 出口 (exports), respectively. We control the model to generate the specific word by selecting the corresponding head and copying its attention weights to the other $H-1$ heads.
In this way, we indeed obtain different translation results with expected translation candidates (see three translation outputs in Table~\ref{table:controlling}). 
\end{CJK}

\begin{CJK}{UTF8}{gbsn}
    \begin{table}\footnotesize
        \centering
        \begin{tabular}{p{1.4cm}|p{5.6cm}}
            Source & 他 说 , 去年 九月 \textbf{以来} ,  \textbf{出口} \textbf{下降} 导致 印度 经济 恶化 。 \\ \hline
            Reference & he said : the drop in exports has caused india 's economy deterioration since september last year . \\ \hline
            Translated & he said : \\ \hline
            以来\,\,\,\,\,\,\,\,\,\,\,\,\,\,\, (since) & he said : \textbf{since} september last year , the decline in exports has led to a deterioration in india 's economy . \\ \hline
            下降\,\,\,\,\,\,\,\,\,\,\,\,\,\,\, (decline) & he said : \textbf{the decline} in exports has led to a deterioration in india 's economy since september last year . \\ \hline
            出口\,\,\,\,\,\,\,\,\,\,\,\,\,\,\, (exports) & he said : \textbf{exports} have declined since september last year , causing india 's economy to deteriorate . \\
        \end{tabular}
        \caption{The model has translated ``he said :" and waits for the following context. Different heads (refered to different candidates) determine different following generation.}
        \label{table:controlling}
    \end{table}
\end{CJK}

Intuitively, we can utilize these characteristics to generate diverse translations by picking different candidates to change the word choices or the sentence structure. More importantly, the diversity is from an interpretable mechanism rather than an abstract latent variable like previous works.

\begin{algorithm}[t]
\footnotesize
\caption{Sample Policy}
\label{alg:sample-algorithm}
    \begin{flushleft}
        \hspace*{0.02in} {\bf Input:} The source sentence length $T$, a hyper parameters $K$, the head number $H$, a counting array $[n_0, ..., n_i, ..., n_{T-1}]$\\
        \hspace*{0.02in} {\bf Output:} Adjusted attention
    \end{flushleft}
    \begin{algorithmic}[1]
    \FOR {$t$ in decoding timesteps}
        \FOR {$i$ in range($T$)}
            \STATE $n_i = 0$
        \ENDFOR
        \STATE calculate $att_{it}^h,\quad i\in[0,T), h\in[0,H)$
        \FOR {$h$ in range($H$)}
            \STATE $candidate_{t}^{h} = \mathop{\arg\max}_i att_{it}^h$
            \STATE $n_{candidate_{t}^{h}} += 1$
        \ENDFOR
        \IF {$max(n) \le K$}
            \STATE $\text{head} = sample[0, H)$
            \FORALL {$h$} 
                \STATE $att_{it}^h = att_{it}^{\text{head}}$ 
            \ENDFOR
        \ENDIF
    \ENDFOR
    \end{algorithmic}
\end{algorithm}

\section{Diversity-Encouraged Generation}
\label{sec:methods}

Since we have confirmed multiple head alignments can be utilized, it is natural for us to sample different heads at every timestep, so that diverse word candidates can be generated. However, we found that it will badly harm the translation quality if sampling everywhere. So we propose a sample policy to balance quality and diversity. 

As stated in Algorithm~\ref{alg:sample-algorithm}, at every decoding step $t$, we denote $att_{it}^h$ as the attention of $head_h$ from target side hidden state $s_t$ to source side word $src_i$. The most possible candidate for $head_h$ next step is :
\begin{equation}
    candidate_{t}^{h} = \mathop{\arg\max}_{i} att_{it}^h
    \label{eq:candidate}
\end{equation}
We denote an array of $[n_0, ..., n_i, ..., n_{T-1}]$ as the number of times that $word_i$ is chosen from equation~\ref{eq:candidate}, where \textit{T} is the length of source sentence. Obviously:
\begin{equation}
    \sum\limits_{i=0}^{T-1} n_i = H
\end{equation}
where $H$ is the number of heads. Diverse translations are generated when multiple candidates are offered. In other words, not all heads focus on the same source word. Therefore, we define a \emph{confusing} condition when:
\begin{equation}
    max(n_i) \le K
\end{equation}
where $K$ is a hyper-parameter. \emph{Confusing} condition means ``referred words" are disperse and multiple candidates can be accepted. Under \emph{confusing} condition, we sample one of the heads as attention and force other heads to be the same. Otherwise, the decoding step remains unchanged. If $K=0$, the model is the same as the original version. If $K=H$, the model samples at every step. For the sake of balance of quality and diversity, we may choose different $K$ in different conditions. We do the whole decoding for $M$ times and pick the most possible output in the beam every time. In this paper, we let $M=5$. 

Our another contribution is adopting the method with back-translation technique as data augmentation. Back-translation has been proved helpful for neural machine translation \cite{sennrich2016Monolingual,Poncelas2018InvestigatingBI}. However, the lack of diversity restricts its effect \cite{Edunov2018UnderstandingBA}. We provide a new scheme for back-translation with diverse corpus generated by our method and gain improvement. 

\section{Experiment}
\label{sec:experiment}
Our translation experiments include two parts: diverse translation and diverse back-translation. In addition, a conversation response generation experiment is also performed as auxiliary evidence. 

\subsection{Setup}
\label{subsec:dataset}
\textbf{Datasets} We choose five datasets as our experiment corpus.

\begin{itemize}
    \item NIST Chinese-to-English (NIST Zh-En). The training data consists of 1.34 million sentence pairs extracted from LDC corpus. We use MT03 as the development set, MT04, MT05, MT06 as the test sets.

    \item WMT14 English-to-German (WMT En-De). The training data consists of 4.5 million sentence pairs from WMT14 news translation task. We use newstest2013 as the development set and newstest2014 as the test set.
    
    \item WMT16 English-to-Romanian (WMT En-Ro). The training data consists of 0.6 million sentence pairs from WMT16 news translation task. We use newstest2015 as the development set and newstest2016 as the test set.
    
    \item Monolingual English corpus (for back-translation) from IWSLT17 Chinese-to-English (IWSLT Zh-En). The training data consists of 0.2 million sentences from IWSlT17 spoken language translation task. We used dev2010 and tst2010 as the development set and tst2011 as the test set.
    
    \item Short Text Conversation (STC) \cite{Shang2015NeuralRM}. The corpus contains about 4.4 million Chinese post-response sentence pairs crawled from Weibo, built for single turn conversation tasks. We remove sentence pairs that are exactly the same between two sides. For the test set, we extract 3000 post sentences that have 10 responses in the corpus, forming 10 references. The develop set is made up similarly.
\end{itemize}

For NIST Zh-En, we use BPE \cite{sennrich2016bpe} with 30K merge operations on both sides. For En-De and En-Ro, we also apply BPE to segment sentences and limit the vocabulary size to 32K. We filter out sentence pairs whose source or target side contains more than 100 words for Zh-En and En-Ro sets. For STC corpus, we also apply BPE and keep a vocabulary size to 36K. All the out-of-vocabulary words are mapped to a distinct token $<$UNK$>$.

\paragraph{Experiment Settings} Without extra statement, we follow the Transformer base v1 settings\footnote{\url{https://github.com/tensorflow/tensor2tensor/blob/v1.3.0/tensor2tensor/models/transformer.py}}, with 6 layers in encoder and 2 layers in decoder\footnote{We check different decoder layer numbers settings and find less-decoder-layers Transformer shows comparable performance with original six-layers-decoder Transformer while it is much easier to manipulate and faster to decode. The diversity enhancement is also more significant. We present detailed research in Appendix, which is highly recommended to refer to.}, 512 hidden units, 8 heads in multi-head attention and 2048 hidden units in feed-forward layers. Parameters are optimized using Adam optimizer \cite{Kingma2015AdamAM}, with $\beta_1 = 0.9$, $\beta_2 = 0.98$, and $\epsilon = 10^{-9}$. The learning rate is scheduled according to the method proposed in \citet{Vaswani2017AttentionIA}, with $warmup\_steps = 8000$. Label smoothing \cite{Szegedy2016RethinkingTI} of $value=0.1$ is also adopted. For $K$, we do not observe diversity enhancement when $K$ is too small like $K=1,2$. And conditions of $K=6,7$ are very similar to $K=8$. Hence we use $K=3,4,5,8$ as comparisons.

\begin{table*}[t]\footnotesize
    \centering
    \begin{tabular}{lccccccccccc}
         & \multicolumn{2}{c}{MT03 (dev)} & \multicolumn{2}{c}{MT04} & \multicolumn{2}{c}{MT05} & \multicolumn{2}{c}{MT06} & \multicolumn{3}{c}{Average}\\ \cline{1-12}
        Model & rfb$\uparrow$ & pwb$\downarrow$ & rfb$\uparrow$ & pwb$\downarrow$ & rfb$\uparrow$ & pwb$\downarrow$ & rfb$\uparrow$ & pwb$\downarrow$ & rfb$\uparrow$ & pwb$\downarrow$ & DEQ$\uparrow$  \\ \hline
        Baseline & 45.64 & 84.63 & 47.25 & 84.62 & 43.45 & 84.78 & 42.26 & 82.46 & 44.32 & 83.95 & - \\
        Multinomial Sampling & 21.75 & 11.29 & 22.19 & 11.42 & 20.54 & 11.08 & 19.12 & 9.67 & 20.62 & 10.72 & 3.09 \\
        \cite{Li2016ASF} & 44.63 & 80.92 & 45.81 & 81.33 & 42.86 & 81.28 & 40.87 & 78.11 & 43.18 & 80.24 & 3.25 \\
        \cite{Vijayakumar2018DiverseBS} & 40.38 & 59.55 & 41.99 & 60.11 & 39.46 & 59.56 & 37.28 & 54.54 & 39.58 & 58.07 & 5.46 \\
        \cite{Shen2019MixtureMF} & 40.59 & 62.24 & 41.55 & 62.68 & 38.51 & 61.37 & 35.57 & 58.04 & 38.54 & 60.70 & 4.02 \\ \hline
        Sample $K=3$ & \textbf{43.73} & \textbf{66.48} & \textbf{45.38} & \textbf{67.82} & \textbf{42.43} & \textbf{65.80} & \textbf{40.18} & \textbf{64.93} & \textbf{42.66} & \textbf{66.18} & \textbf{10.70} \\
        Sample $K=4$ & 40.88 & 51.26 & 42.50 & 53.63 & 39.18 & 51.07 & 37.73 & 50.28 & 39.80 & 51.66 & 7.14 \\
        Sample $K=5$ & 38.60 & 43.64 & 40.21 & 45.69 & 37.05 & 43.14 & 35.45 & 42.38 & 37.57 & 43.74 & 5.96 \\
        Sample $K=8$ & 36.68 & 38.29 & 38.03 & 40.02 & 34.65 & 37.30 & 32.93 & 36.15 & 35.20 & 37.82 & 5.06 \\
    \end{tabular}
    \caption{Pair-wise BLEU and Reference BLEU in Zh2En experiments.}
    \label{table:diversity-zh2en}
\end{table*}

\begin{table}[t] \footnotesize
    \centering
    \begin{tabular}{lccc}
        Model & rfb$\uparrow$ & pwb$\downarrow$ & DEQ$\uparrow$\\ \hline
        Baseline & 26.31 & 80.41 & - \\
        Multinomial Sampling & 11.99 & 12.84 & 4.72 \\
        \cite{Li2016ASF} & 25.27 & 78.57 & 1.77\\ 
        \cite{Vijayakumar2018DiverseBS} & 23.27 & 66.13 & 4.70\\ 
        \cite{Shen2019MixtureMF} & 23.22 & 68.03 & 4.01\\ \hline
        Sample $K=3$ & 25.62 & 78.96 & 2.10\\
        Sample $K=4$ & \textbf{24.26} & \textbf{62.04} & \textbf{8.96} \\
        Sample $K=5$ & 22.62 & 50.14 & 8.20\\
        Sample $K=8$ & 19.76 & 38.36 & 6.42
    \end{tabular}
    \caption{Pair-wise BLEU and Reference BLEU in En2De experiments.}
    \label{table:diversity-en2de}
\end{table}

\begin{table}[t]
\footnotesize
    \centering
    \begin{tabular}{lccc}
        Model & rfb$\uparrow$ & pwb$\downarrow$ & DEQ$\uparrow$ \\ \hline
        Baseline & 31.76 & 81.29 & - \\
        Multinomial Sampling & 18.85 & 20.82 & 4.68\\
        \cite{Li2016ASF} & 31.02 & 78.42 & 3.88 \\ 
        \cite{Vijayakumar2018DiverseBS} & 28.91 & 69.67 & 4.08\\ 
        \cite{Shen2019MixtureMF} & 31.07 & 85.71 & -6.04\\\hline
        Sample $K=3$ & 31.33 & 82.41 & -2.60 \\
        Sample $K=4$ & \textbf{30.06} & \textbf{71.12} & \textbf{5.98} \\
        Sample $K=5$ & 27.89 & 59.42 & 5.65\\
        Sample $K=8$ & 26.43 & 50.56 & 5.77
    \end{tabular}
    \caption{Pair-wise BLEU and Reference BLEU in En2Ro experiments.}
    \label{table:diversity-en2ro}
\end{table}

\subsection{Diverse Translation}

\paragraph{Comparing Objects} We compare our models with original beam search (as Baseline) and sampling from the probability distribution (as Multinomial Sampling). Besides, we compare our methods with a few previous works:

\begin{itemize}
    \item \citet{Li2016ASF} propose a decoding trick to penalize hypotheses that are siblings (expansions of the same parent node) in the beam search to increase the translation diversity.
    
    \item \citet{Vijayakumar2018DiverseBS} add a regularization item in beam search to penalize the same word generation.
    
    
    \item \citet{Shen2019MixtureMF} and \citet{He2018SequenceTS} use multiple decoders as mixture of experts to increase diversity by manipulating latent variables. Considering their similarity, we choose \citet{Shen2019MixtureMF} since they report better results. We re-implement the model with Transformer architecture and choose the hMup (online-shared) version since the authors recommend it.
\end{itemize}

\begin{figure}
    \centering                 
    \includegraphics[scale=0.4]{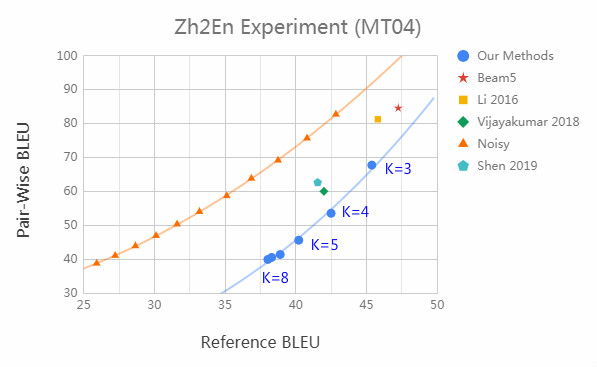}   
    \caption{Pair-wise BLEU with reference BLEU in Zh2En Experiments (MT04). The bottom right corner means the best result. All previous work including noisy sets lie on the top left of the curve of $K$. }
    \label{fig:curve}
\end{figure}
\begin{CJK}{UTF8}{gbsn}
    \begin{table*}\scriptsize
        \centering
        \begin{tabular}{lp{15cm}}
            Input & 两 个 主角 — — 朝鲜 和 美国 都 没有 表现 出 让步 , 双方 的 基本 立场 也 都 没有 松动 。 \\ \hline
            Beam & 1. the two leading characters -- the dprk and the united states -- did not make any concessions , and the basic positions of both sides were not relaxed . \\
             & 2. the two leading characters -- the dprk and the united states -- did not make any concessions , and the basic positions of both sides were not loosened . \\
             & 3. the two leading characters -- the dprk and the united states -- did not make any concessions , and both sides did not relax their basic positions . \\
             & 4. the two leading characters -- the dprk and the united states -- did not make any concessions . both sides did not relax their basic positions . \\
             & 5. the two leading characters -- the dprk and the united states -- did not make any concessions , and both sides ' basic positions were not relaxed . \\ \hline
            K=4 & 1. the two leading characters -- the dprk and the united states -- did not make any concessions . both sides ' basic positions were not relaxed . \\
             & 2. neither the dprk nor the united states has made any concessions . both sides have not relaxed their basic positions . \\
             & 3. the two leading roles -- the dprk and the united states -- have made no concessions , and neither have they relaxed their basic positions . \\
             & 4. neither the dprk nor the united states -- the two leading characters -- did make any concessions , and the basic positions of both sides were not relaxed .\\
             & 5. neither the democratic people 's republic of korea and the united states have made any concessions , and the basic positions of both sides have not been relaxed .
        \end{tabular}
        \caption{One case. Our method shows obviously more diversity compared with beam search. For more cases, see Appendix.}
        \label{table:cases}
    \end{table*}
\end{CJK}

\paragraph{Metrics} We evaluate our method from both diversity and quality. For diversity, we adopt average pair-wise BLEU of outputs (denoted as \emph{pwb}) to measure the difference among translations like previous work. For quality, we use BLEU with the references (denoted as \emph{rfb}). 
Lower \emph{pwb} and higher \emph{rfb} mean better results. In this paper, the reference BLEU of Baseline is the highest score in the beam while the other groups take the average reference BLEU of M outputs. And to synthetically evaluate the performance, we propose an overall index: Diversity Enhancement per Quality (denoted as \emph{DEQ}). Specifically:

\begin{equation}
     \text{DEQ} = \frac{\text{pwb}^*-\text{pwb}}{\text{rfb}^*-\text{rfb}}
\end{equation}

where pwb and pwb* refer to pair-wise BLEU of the evaluated system and baseline respectively, rfb and rfb* refer to reference BLEU of the evaluated system and baseline respectively. It measures how much diversity can be produced per quality drop.

\paragraph{Results} From Table~\ref{table:diversity-zh2en}, in Zh2En experiment, we can see that traditional beam search translations severely lack diversity while multinomial sampling extremely harms translation quality. \citet{Li2016ASF} bring very limited enhancement in diversity, failing to achieve the goal. \citet{Vijayakumar2018DiverseBS} and \citet{Shen2019MixtureMF} show the ability to produce diversity, but our method ($K=4$) attains more significant diversity as well as better quality comparing with them. And $K=3$ achieve the highest $DEQ$, gaining the most satisfactory result. Also, unlike \citet{Shen2019MixtureMF}, our work needs no extra training or extra parameters. Furthermore, the diversity can be well interpreted and does not rely on an abstract latent variable. See Table~\ref{table:cases} for a case (more cases are in Appendix). We can reach the similar conclusion in En2De and En2Ro experiments from Table~\ref{table:diversity-en2de}, \ref{table:diversity-en2ro}.


Besides, to exclude the possibility that randomly interfering causes the effect, we compare with the sets with noise. We add noise to the translations of baseline model to generate different outputs. Specifically, for each sentence, we replace one of its words with `$<$UNK$>$' with probability $p$ and randomly swap two words with probability $p$ as well. And we make multiple experiment sets by controlling $p$. See Figure~\ref{fig:curve}, at the same level of pair-wise BLEU, our method maintains much higher reference BLEU, which means our method improves diversity through seeking diverse translations rather than just generating randomly.

For $K$ (see Figure~\ref{fig:curve} again), as expected, as $K$ grows (sample more), the diversity increases (pair-wise BLEU decreases) while the quality decreases (reference BLEU decreases). And previous works all lie on the top left of the curve of $K$. What's more, we can choose different $K$ to diversely balance the diversity and the quality depending on our needs. We make the trade-off more continuous.

Some may not be satisfied with the sacrifice of the reference BLEU. But considering the calculation of BLEU is based on n-gram rather than semantic similarity, we regard it as a normal phenomenon. After all, if we want to obtain sentences with different grammar structure or word order, the overlap of n-gram will inevitably decrease to some extent even the meaning remains the same. Meanwhile, we empirically prove our method maintains a relatively high quality comparing to noisy sets as well as previous works. 

\begin{figure}
    \centering                 
    \includegraphics[scale=0.065]{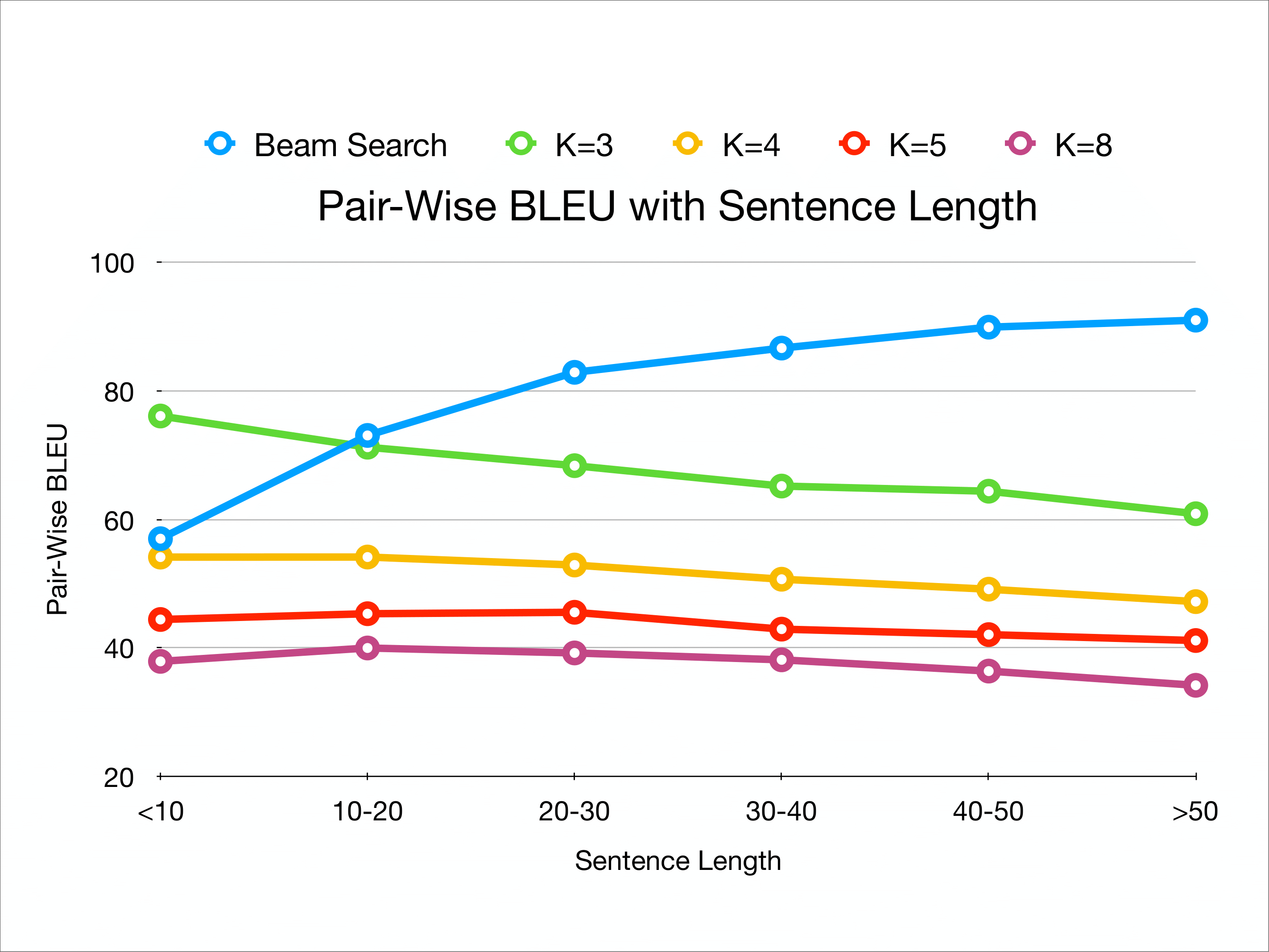}
    \caption{Our method increases diversity (pair-wise BLEU decreases) as the sentences getting longer.}
    \label{fig:pw-bleu}
\end{figure}

We also investigate the effect of sentence length. Theoretically, longer sentences shall have more diversity due to their broader searching space. However, beam search with MAP prefers to abandon different but slightly less possible candidates, making hypotheses lack diversity and are all close to specific translations. Conversely, our method increases diversity as the sentences getting longer (see figure~\ref{fig:pw-bleu}), which conforms to the statistical law.


\subsection{Diverse Back-Translation}

Back-translation has been proved helpful for neural machine translation \cite{sennrich2016Monolingual,Poncelas2018InvestigatingBI}. However, the lack of diversity restricts its effect \cite{Edunov2018UnderstandingBA}. We try to utilize our methods to enhance the translation performance by improving back-translation. 
According to \citet{Edunov2018UnderstandingBA}, unrestricted sampling from the model distribution yields the best performance. 
Therefore, we compare with 1) baseline without utilizing back-translation, denoted as \textit{Baseline}, 2) beam search as back-translation, denoted as \textit{Beam-5}, 3) unrestricted sampling as back-translation, denoted as \textit{Sampling}.

We do experiments under conditions with and without additional monolingual data.

\subsubsection{Self Back-Translation}

Firstly, We focus on the condition where original training data is repeatedly used by back-translation. When translating language pair $f$ to $e$, for each target sentence $e$, we get $M$ translations with a reverse translation model. We combine those translations with $e$ as synthetic sentence pairs and add them to the training data.
As previously stated, we let $M=5$. Experiments are conducted on Zh-En NIST dataset.

See Table~\ref{table:bt-nist-zh2en} and \ref{table:bt-nist-en2zh}, all of our experiment sets report better results, among which, the best set of $K=3$ in Zh2En experiments yields 1.82 improvement and the best set of $K=4$ in En2Zh experiments yields 0.82 improvement.

\subsubsection{Utilizing Additional Monolingual Data}

Secondly, we evaluate our method with additional monolingual data. We select one side of parallel data from IWSLT17 as monolingual data. We use the same method to generate synthetic sentence pairs. Then we train our model on the mixture of original NIST dataset and the synthetic dataset.

See Table~\ref{table:bt-iwslt-zh2en}, experiment results show that our algorithm brings the most significant improvement for translation performance as our work adds generation diversity and maintains the quality simultaneously.

\begin{table}[ht]\footnotesize
    \centering
    \begin{tabular}{lccccc}
        Model & MT03 & MT04 & MT05 & MT06 & Average \\ \hline
        Baseline & 45.64 & 47.25 & 43.45 & 42.26 & 44.32 \\
        Beam-5 & 46.31 & 47.26 & 44.87 & 43.43 & 45.19 \\
        Sampling & 47.03 & 47.96 & 45.72 & 44.06 & 45.91 \\ \hline
        $K=3$ & 47.24 & 48.24 & \textbf{45.70} & \textbf{44.48} & \textbf{46.14} \\
        $K=4$ & \textbf{47.39} & 47.93 & 45.38 & 43.98 & 45.76 \\
        $K=5$ & 47.31 & \textbf{48.31} & 45.34 & 43.95 & 45.87 \\
        $K=8$ & 47.15 & 48.15 & 45.69 & 43.95 & 45.93 \\
    \end{tabular}
    \caption{Zh2En translation experiments with back-translation of original training data.}
    \label{table:bt-nist-zh2en}
\end{table}

\begin{table}[ht]\footnotesize
    \centering
    \begin{tabular}{lccccc}
        Model & MT03 & MT04 & MT05 & MT06 & Average\\ \hline
        Baseline & 22.75 & 22.33 & 20.35 & 21.35 & 21.34 \\
        Beam-5 & 23.73 & 21.69 & 20.61 & 22.33 & 21.54 \\
        Sampling & 23.69 & 22.78 & 20.85 & 22.34 & 21.99 \\ \hline
        $K=3$ & \textbf{24.21} & 22.23 & 20.65 & \textbf{22.52} & 21.80 \\
        $K=4$ & 24.01 & \textbf{23.15} & \textbf{21.04} & 22.30 & \textbf{22.16} \\
        $K=5$ & 23.76 & 21.93 & 20.57 & 22.50 & 21.67 \\
        $K=8$ & 23.93 & 21.66 & 20.72 & 22.23 & 21.54 \\
    \end{tabular}
    \caption{En2Zh translation experiments with back-translation of original training data.}
    \label{table:bt-nist-en2zh}
\end{table}

\begin{table}[ht]\footnotesize
    \centering
    \begin{tabular}{lc}
        Model & Zh2En \\ \hline
        Baseline & 9.18 \\
        Beam-5 & 13.06 \\
        Sampling & 13.38 \\ 
        \hline
        $K=3$ & \textbf{14.03} \\
        $K=4$ & 13.76 \\
        $K=5$ & 13.66 \\
        $K=8$ & 13.76 \\
    \end{tabular}
    \caption{Zh2En translation experiments with back-translation with additional monolingual data.}
    \label{table:bt-iwslt-zh2en}
\end{table}


\subsection{Conversation Response Generation}

Responses generated by neural conversational models tend to lack informativeness and diversity \cite{Li2016ADO,Shao2017GeneratingHA,Baheti2018GeneratingMI,Zhang2018GeneratingIA}. Therefore, we try to ease this issue by utilizing our method on Conversation Response Generation tasks. Still, we perform decoding for $M$ times and pick the $N^{th}$ output in the beam for the $N^{th}$ group ($N\in[1,M]$).

\paragraph{Metrics} Since the responses of human conversation can be pretty subjective, which is hard to evaluate automatically. Hence, except for reference BLEU, we also measure the response quality by human evaluation through three indexes: relevance, fluency and informativeness. Relevance reveals how much the responses match the expectation of the question. Fluency means to what extent the translation is well-formed grammatically. Both of them are scored from 1 to 5. Informativeness measures the degree of meaningfulness. We classify responses into two groups, informative and uninformative. Uninformative means the safe answer like ``I don't know" or simply copying from the original post. We then calculate the proportion of the informative groups. For diversity, pair-wise BLEU maintains used. 

\paragraph{Results} In Table~\ref{table:conversation}, we compare our method with basic Seq2Seq model and \citet{Li2016ADO}, which use Maximum Mutual Information (MMI) as the objective function (MMI-antiLM version). On one hand, our method achieves significant improvement in generation diversity. On the other hand, the quality including relevance, fluency and informativeness all rise to some degree. After looking into cases, we suppose it is because original Seq2Seq model tends to generate safe outputs like ``I don't know" or simply copying from the source side. In contrast, our method brings in randomness, reaching broader generation space.

\begin{table}\footnotesize
    \centering
    \begin{tabular}{lccccc}
        & BLEU $\uparrow$ & Rel $\uparrow$ & Flu $\uparrow$ & Inf $\uparrow$ & Div $\downarrow$ \\ \hline
        Baseline & 13.06 & 2.45 & 4.72 & 0.604 & 52.94 \\
        MMI & 13.39 & 2.58 & 4.45 & 0.639 & 45.42 \\ \hline
        $K=3$ & \textbf{13.48} & \textbf{2.63} & \textbf{4.76} & \textbf{0.678} & \textbf{39.82} \\
        $K=8$ & 12.73 & 2.53 & 4.67 & 0.652 & 27.73 \\
    \end{tabular}
    \caption{Conversation Response Generation experiment results on STC dataset.}
    \label{table:conversation}
\end{table}




\section{Related Work}


Lack of diversity has been a disturbing problem for neural machine translation. In recent years, a few works put forward some related methods. \citet{Li2016ASF} proposes a decoding trick to penalize hypotheses that are siblings (expansions of the same parent node) in the beam search to increase the translation diversity. \citet{Vijayakumar2018DiverseBS} adds a regularization item in beam search to penalize the same word generation. \citet{He2018SequenceTS} and \citet{Shen2019MixtureMF} use multiple decoders as different components, trying to control the generation by different latent variables. Basically, there are two categories: either to add diverse regularization in beam search or to utilize latent variables. Our method achieves better results than both two categories. And specifically, compared with the latter class, our work needs no extra training or extra parameters. Besides, it is hard to tell what the latent variables exactly represent and why they differ while our method shows a clear explanation that heads align to word candidates.

Apart from machine translation, there are also other works concerning generation diversity, including 
Visual Question Generation \cite{Jain2017CreativityGD}, Conversational Response Generation \cite{Li2016ADO,Shao2017GeneratingHA,Baheti2018GeneratingMI,Zhang2018GeneratingIA}, Paraphrase \cite{Gupta2018ADG,Xu2018DPAGEDP}, Summarization \cite{Nema2017DiversityDA} and Text Generation \cite{Guu2018GeneratingSB,Xu2018DiversityPromotingGA}.

As for multi-head attention, \citet{Strubell2018LinguisticallyInformedSF} employ different heads to capture different linguistic features. \citet{Tu2018MultiHeadAW} introduce disagreement regularization to encourage diversity among attention heads. \citet{Li2019InformationAF} propose to aggregate information captured by different heads. \citet{Yang2019ConvolutionalSN} model the interactions among attention heads. \citet{Raganato2018AnAO} do an analysis of encoder representation and find there exists dependency relations, syntactic and semantic connections across layers.

\section{Conclusion}
In this paper, we discover an internal characteristic of Transformer encoder-decoder multi-head attention that each head aligns to a source word which is a possible candidate to be translated. We take advantage of this phenomenon to generate diverse translations by manipulating heads in particular conditions. Experiments show that our algorithm outperforms previous work and obtain the most satisfactory result of quality and diversity. Besides, the multiple trade-off setting can be adopted diversely depending on different needs. Finally, applications on back-translation as data augmentation and conversation response significantly improve the performance, proving our method effective.

\section{Acknowledgement}

Shujian Huang is the corresponding author. This work is supported by the National Key R\&D Program of China (No. U1836221, 61772261), the Jiangsu Provincial Research Foundation for Basic Research (No. BK20170074).

\bibliography{aaai}
\bibliographystyle{aaai}

\clearpage

\appendix

\section{Less-decoder-layers Transformer works fine as well}
\label{sec:less-layer}

\begin{figure}[ht]
    \centering                 
    \includegraphics[scale=0.55]{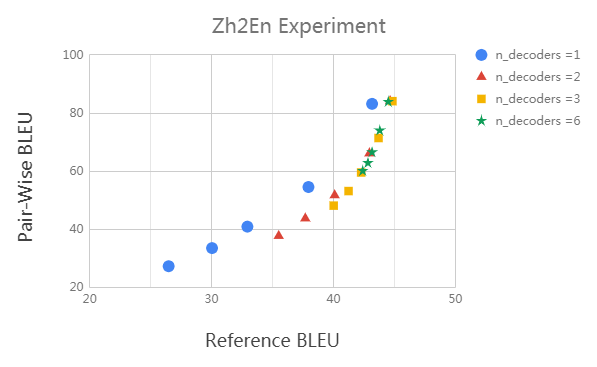}   
    \caption{Pair-wise BLEU with Reference BLEU in Zh2En experiments of different decoder layer numbers and different K sets.}
    \label{fig:layers-comparisons-for-diversity}
\end{figure}

We check the different experimental groups of different decoder layer numbers to see whether the alignments exist widely. It turns out that less-decoder-layers Transformer (six-layers encoder unchanged) has this property as well. And owing to its shallower structure, we find the phenomenon more common and obvious. Simultaneously, it is much easier to control the word candidates by picking different heads from the experiment. For diversity, see Table~\ref{table:layers-comparisons-for-diversity} and Figure~\ref{fig:layers-comparisons-for-diversity}.

As for the performance, to our surprise, less-decoder-layers Transformer performs comparably with six-decoder-layers Transformer (see Table~\ref{table:layers-comparisons-for-bleu} \ref{table:layers-comparisons-en2de&en2ro}). What's more, less decoder layers bring higher speed (see Table~\ref{table:layers-comparisons-for-bleu}).

Thus, for the sake of diversity significance and decoding speed, we choose the two-layers-decoder Transformer structure as our default setting unless stated otherwise.

\section{More cases}
\label{appendix:more-cases}

We select some typical cases of diverse translation from Zh2En experiments (see Table~\ref{table:more-cases}).


\newpage

\begin{table}[ht] \footnotesize
    \centering
    \begin{tabular}{ccccccc}
        n\_d & MT03 & MT04 & MT05 & MT06 & Average & Speed \\ \hline
        6 & 45.83 & 46.66 & 43.36 & 42.17 & 44.06 & 1 X   \\ \hline
        1 & 44.55 & 45.73 & 41.86 & 40.53 & 42.71 & 2.79 X  \\
        2 & \textbf{45.64} & \textbf{47.25} & \textbf{43.45} & \textbf{42.26} & \textbf{44.32} & 2.34 X  \\
        3 & \textbf{45.85} & \textbf{46.99} & \textbf{43.74} & \textbf{42.77} & \textbf{44.50} & 1.77 X  \\
        4 & 45.29 & 46.77 & 43.71 & 42.09 & 44.19 & 1.58 X  \\
        5 & 45.27 & 46.04 & 43.33 & 43.06 & 44.14 & 1.26 X  \\
        8 & 44.81 & 46.11 & 42.24 & 42.15 & 43.50 & 0.69 X \\
        12 & 45.34 & 45.49 & 43.09 & 42.41 & 43.66 & 0.67 X \\
    \end{tabular}
    \caption{BLEU and speed comparisons between different decoder layer numbers in Zh2En Experiments}
    \label{table:layers-comparisons-for-bleu}
\end{table}

\begin{table}[ht]
    \centering
    \begin{tabular}{lcc}
        n\_decoders & En2De & En2Ro  \\ \hline
        two & 26.31 & 31.76 \\
        six & 26.70 & 31.86
    \end{tabular}
    \caption{BLEU comparisons between different decoder layer numbers in En2De and En2Ro Experiments}
    \label{table:layers-comparisons-en2de&en2ro}
\end{table}

\begin{table*}[ht]
    \centering
    \begin{tabular}{ll|cccccccccc} 
         & & \multicolumn{2}{c}{MT03 (dev)} & \multicolumn{2}{c}{MT04} & \multicolumn{2}{c}{MT05} & \multicolumn{2}{c}{MT06} & \multicolumn{2}{c}{Average}\\ \cline{3-12}
        n\_d & Sample K & rfb$\uparrow$ & pwb$\downarrow$ & rfb$\uparrow$ & pwb$\downarrow$ & rfb$\uparrow$ & pwb$\downarrow$ & rfb$\uparrow$ & pwb$\downarrow$ & rfb$\uparrow$ & pwb$\downarrow$  \\ \hline
        \multirow{5}*{1} & $Baseline$ & 44.55 & 83.38 & 45.73 & 83.68 & 41.86 & 84.28 & 40.53 & 81.32 & 42.71 & 83.09 \\
        & $K=3$ & 39.15 & 54.91 & 40.18 & 56.29 & 37.02 & 53.58 & 35.47 & 53.44 & 37.56 & 54.44 \\
        & $K=4$ & 33.86 & 40.96 & 35.39 & 43.09 & 32.11 & 40.07 & 30.38 & 39.58 & 32.63 & 40.91 \\
        & $K=5$ & 30.89 & 33.31 & 32.37 & 35.42 & 29.26 & 33.01 & 27.66 & 32.43 & 29.76 & 33.62 \\
        & $K=8$ & 26.94 & 26.63 & 28.99 & 28.97 & 25.57 & 26.99 & 24.41 & 26.62 & 26.32 & 27.53 \\ \hline
        \multirow{5}*{2} & $Baseline$ & 45.64 & 84.63 & 47.25 & 84.62 & 43.45 & 84.78 & 42.26 & 82.46 & 44.32 & 83.95 \\
        & $K=3$ & 43.73 & 66.48 & 45.38 & 67.82 & 42.43 & 65.80 & 40.18 & 64.93 & 42.66 & 66.18 \\
        & $K=4$ & 40.88 & 51.26 & 42.50 & 53.63 & 39.18 & 51.07 & 37.73 & 50.28 & 39.80 & 51.66 \\
        & $K=5$ & 38.60 & 43.64 & 40.21 & 45.69 & 37.05 & 43.14 & 35.45 & 42.38 & 37.57 & 43.74 \\
        & $K=8$ & 36.68 & 38.29 & 38.03 & 40.02 & 34.65 & 37.30 & 32.93 & 36.15 & 35.20 & 37.82 \\ \hline
        \multirow{5}*{3} & $Baseline$ & 45.85 & 84.28 & 46.99 & 84.72 & 43.74 & 85.12 & 42.77 & 82.24 & 44.5 & 84.03 \\
        & $K=3$ & 44.39 & 72.03 & 46.02 & 72.81 & 42.96 & 70.74 & 41.45 & 70.06 & 43.48 & 71.20 \\
        & $K=4$ & 42.91 & 59.93 & 44.59 & 61.15 & 41.50 & 58.77 & 40.07 & 58.34 & 42.05 & 59.42 \\
        & $K=5$ & 41.90 & 53.15 & 43.64 & 54.92 & 40.36 & 52.71 & 39.07 & 51.87 & 41.02 & 53.17 \\
        & $K=8$ & 40.92 & 48.34 & 42.23 & 50.03 & 38.93 & 47.44 & 38.00 & 46.77 & 39.72 & 48.08 \\ \hline
        \multirow{5}*{4} & $Baseline$ & 45.29 & 84.35 & 46.77 & 84.62 & 43.71 & 84.70 & 42.09 & 82.25 & 44.19 & 83.86 \\
        & $K=3$ & 44.25 & 72.44 & 46.05 & 74.57 & 42.53 & 72.42 & 41.75 & 71.64 & 43.44 & 72.88 \\
        & $K=4$ & 43.35 & 62.50 & 44.95 & 64.43 & 41.71 & 62.64 & 40.81 & 61.43 & 42.49 & 62.83 \\
        & $K=5$ & 42.44 & 57.72 & 44.33 & 59.36 & 40.93 & 57.40 & 40.24 & 55.91 & 41.83 & 57.56 \\
        & $K=8$ & 41.43 & 52.84 & 43.31 & 55.15 & 39.92 & 53.29 & 39.32 & 52.05 & 40.85 & 53.50 \\ \hline
        \multirow{5}*{5} & $Baseline$ & 45.27 & 84.39 & 46.04 & 84.74 & 43.33 & 84.92 & 43.06 & 82.47 & 44.14 & 84.04 \\
        & $K=3$ & 44.41 & 74.45 & 45.58 & 75.84 & 42.66 & 73.61 & 42.32 & 72.71 & 43.52 & 74.05 \\
        & $K=4$ & 43.61 & 65.59 & 44.77 & 67.01 & 42.04 & 64.83 & 41.73 & 64.03 & 42.85 & 65.29 \\
        & $K=5$ & 42.86 & 60.87 & 44.34 & 62.54 & 41.16 & 60.28 & 41.12 & 59.46 & 42.21 & 60.76 \\
        & $K=8$ & 42.21 & 57.16 & 43.88 & 59.14 & 40.55 & 56.61 & 40.42 & 55.49 & 41.62 & 57.08 \\ \hline
        \multirow{5}*{6} & $Baseline$ & 45.83 & 84.24 & 46.66 & 84.43 & 43.36 & 84.71 & 42.17 & 82.16 & 44.06 & 83.77 \\
        & $K=3$ & 44.76 & 74.36 & 45.99 & 75.42 & 42.85 & 73.65 & 41.61 & 72.65 & 43.48 & 73.91 \\
        & $K=4$ & 43.87 & 66.50 & 45.50 & 68.26 & 42.08 & 65.89 & 41.25 & 65.51 & 42.94 & 66.55 \\
        & $K=5$ & 43.65 & 62.53 & 45.13 & 64.89 & 41.73 & 62.32 & 40.83 & 61.61 & 42.56 & 62.94 \\
        & $K=8$ & 42.98 & 59.58 & 44.61 & 62.22 & 41.31 & 59.61 & 40.69 & 59.05 & 42.20 & 60.29 \\ \hline
        \multirow{5}*{8} & $Baseline$ & 44.81 & 84.55 & 46.11 & 84.77 & 42.24 & 85.11 & 42.15 & 82.70 & 43.5 & 84.19 \\
        & $K=3$ & 43.80 & 73.45 & 45.67 & 75.91 & 41.81 & 73.38 & 41.58 & 73.10 & 43.02 & 74.13 \\
        & $K=4$ & 43.53 & 67.72 & 45.43 & 70.30 & 41.35 & 67.54 & 41.39 & 67.31 & 42.72 & 68.38 \\
        & $K=5$ & 43.59 & 65.29 & 45.09 & 68.05 & 41.25 & 64.65 & 41.10 & 64.12 & 42.48 & 65.61 \\
        & $K=8$ & 42.84 & 63.57 & 45.18 & 66.64 & 41.21 & 63.21 & 40.85 & 62.92 & 42.41 & 64.26 \\ \hline
        \multirow{5}*{12} & $Baseline$ & 45.34 & 84.51 & 45.49 & 84.34 & 43.09 & 84.86 & 42.41 & 82.15 & 43.66 & 83.78 \\
        & $K=3$ & 44.84 & 75.90 & 45.33 & 77.66 & 42.47 & 76.00 & 42.32 & 74.97 & 43.37 & 76.21 \\
        & $K=4$ & 44.77 & 72.38 & 45.21 & 73.85 & 42.33 & 72.37 & 42.02 & 71.21 & 43.19 & 72.48 \\
        & $K=5$ & 44.57 & 70.53 & 45.19 & 72.51 & 42.24 & 70.72 & 42.06 & 69.71 & 43.16 & 70.98 \\
        & $K=8$ & 44.73 & 70.43 & 45.18 & 72.40 & 42.30 & 69.95 & 41.91 & 68.85 & 43.13 & 70.40 \\

    \end{tabular}
    \caption{Pair-wise BLEU and Reference BLEU in Zh2En experiments of different decoder layer numbers and different $K$ sets.}
    \label{table:layers-comparisons-for-diversity}
\end{table*}

\begin{CJK}{UTF8}{gbsn}
    \begin{table*}[t]\footnotesize
        \centering
        \begin{tabular}{lp{14cm}}
            Input & 只有 走 以 最 有效 利用 资源 和 保护 环境 为 基础 的 循环 经济 之 路 , 可 持续 发展 才 能 得到 实现 。 \\ \hline
            Reference & the only route to sustainable development is a recycle economy , which is based on the foundation of environmental protection and makes the most efficient use of resources . \\ \hline
            K=4 & 1. sustainable development is realized . only by taking the road of a cycle economy based on the availability of resources and environmental protection . \\
            & 2. only by taking the cycle of the economy based on the most effective use of resources and environmental protection can we achieve sustainable development . \\
            & 3. it is only through following the path of a cycle economy based on the most effective use of resources and environmental protection that sustainable development can be realized . \\
            & 4. sustainable development can only be realized if we take the most effective means of utilizing resources and protecting the environment . \\
            & 5. sustainable development can only be realized by taking the road of a cycle economy which is the most effective use of resources and environmental protection . \\ \\ \\
        \end{tabular}
        
        \begin{tabular}{lp{14cm}}
            Input & 由于 外界 猜测 印度 和 巴基斯坦 两 国 领导人 有 可能 在 会议 期间 就 双边 关系 问题 举行 会谈 , 这次 南盟 首脑 会议 格外 引人注目 。 \\ \hline
            Reference & as outsiders are guessing the leaders of india and pakistan might be holding a meeting on their bilateral relation , the south asia summit becomes especially conspicuous . \\ \hline
            K=4 & 1. as the outside world conjecture that leaders of india and pakistan may hold talks on bilateral relations , this summit will attract people 's attention . \\
            & 2. as people have been speculating that india and pakistan may hold talks on issues of bilateral relations during the meeting , the current summit of heads of state has attracted the attention of the outside world . \\
            & 3. as speculations by outsiders that leaders of india and pakistan may hold talks on bilateral relations during the meeting , this summit meeting of the heads of state has attracted much attention . \\
            & 4. as outsiders guessed that leaders of india and pakistan could possibly hold talks on bilateral relations , the summit meeting was particularly eye-catching . \\
            & 5. as outsiders guessed that india and pakistan might hold talks on bilateral relations , this summit was particularly eye-catching . \\ \\ \\
        \end{tabular}
        
        \begin{tabular}{lp{14cm}}
            Input & 2003年 , 空客 公司 的 飞机 交付 量 占 全球 的 份额 跃 升 至 52 \% , 2001年 和 2002年 其 全球 份额 分别 为 38 \% 和 44 \% 。 \\ \hline
            Reference & in 2003 , airbus company 's delivery increased to 52 \% of the global market . its market share for 2001 and 2002 were 38 \% and 44 \% respectively . \\ \hline
            K=4 & 1. air passenger traffic accounted for 52 percent of the global share in 2003 and 38 percent in 2001 and 44 percent in 2002 respectively . \\
            & 2. the share of aircraft delivered by air passenger companies has risen to 52 percent , while the global share in 2001 and 2002 was 38 percent and 44 percent . \\
            & 3. the share of air passenger companies in the world jumped to 52 percent , and in 2001 and 2002 , 38 percent and 44 percent respectively . \\
            & 4. air passenger companies accounted for 52 percent of the world 's share , and 38 percent and 44 percent of the world 's world share in 2001 and 2002 . \\
            & 5. in 2003 , the air passenger companies accounted for 52 percent of the world 's aircraft delivery and their global share was 38 percent in 2001 and 44 percent in 2002 . \\ \\ \\
        \end{tabular}
        
        \begin{tabular}{lp{14cm}}
            Input & 不过 他 认为 , 经过 美军 长期 训练 后 , 伊拉克 部队 的 训练 会 获得 成果 。 \\ \hline
            Reference & but he believes that after being trained for a long time by the us army , iraqi forces will achieve good training results . \\ \hline
            K=4 & 1. he said , however , that after a us military training , the training of iraqi troops will be fruitful . \\
            & 2. he , however , believes that after us training for the us military over a long period of time , iraq 's training will be successful . \\
            & 3. however , he believed that the iraqi troops ' training will be successful after a long period of training us troops . \\
            & 4. he said , however , that iraq 's troops ' training has been successful after the us military has trained for a long time . \\
            & 5. he , however , believes that iraq 's troops ' training will yield results after training for the us military over a long period of time . 
        \end{tabular}
        
        \caption{More Cases}
        \label{table:more-cases}
    \end{table*}
\end{CJK}

\end{document}